\documentclass[final]{cvpr}

\usepackage{times}
\usepackage{epsfig}
\usepackage{graphicx}
\usepackage{amsmath}
\usepackage{amssymb}
\usepackage{subcaption}
\usepackage{algorithm}
\usepackage{algpseudocode}

\usepackage{pifont}
\usepackage[pagebackref=true,breaklinks=true,colorlinks,bookmarks=false]{hyperref}

\def\cvprPaperID{****} 
\def\confYear{CVPR 2021}

\begin{document}

\title{Boosting Co-teaching with Compression Regularization for Label Noise}

\author{Yingyi Chen$^1$ 
\quad Xi Shen$^2$ 
\quad Shell Xu Hu$^3$ 
\quad Johan A.K.~Suykens$^1$ 
\\
{\small 
$^1$ESAT-STADIUS, KU Leuven 
\quad 
$^2$LIGM (UMR 8049), Ecole des Ponts, UPE
\quad 
$^3$Upload AI LLC
\par}
\\
{\tt\small  
\{yingyi.chen,johan.suykens\}@esat.kuleuven.be,
xi.shen@enpc.fr,
shell@uploadai.com}
}


\maketitle

\begin{abstract}
   In this paper, we study the problem of learning image classification models in the presence of label noise. 
   We revisit a simple compression regularization named Nested Dropout~\cite{rippel2014learning}. We find that Nested Dropout~\cite{rippel2014learning}, though originally proposed to perform fast information retrieval and adaptive data compression, can properly regularize a neural network to combat label noise. 
   Moreover, owing to its simplicity, it can be easily combined with Co-teaching~\cite{han2018co} to further boost the performance. 
   
   Our final model remains simple yet effective: it achieves comparable or even better performance than the state-of-the-art approaches on two real-world datasets with label noise which are Clothing1M~\cite{xiao2015learning} and ANIMAL-10N~\cite{song2019selfie}. 
   On Clothing1M~\cite{xiao2015learning}, our approach obtains 74.9\% accuracy which is slightly better than that of DivideMix~\cite{li2020dividemix}. 
   On ANIMAL-10N~\cite{song2019selfie}, we achieve 84.1\% accuracy while the best public result by PLC~\cite{zhang2021learning} is 83.4\%. 
   We hope that our simple approach can be served as a strong baseline for learning with label noise.
   Our implementation is available at \href{https://github.com/yingyichen-cyy/Nested-Co-teaching}{https://github.com/yingyichen-cyy/Nested-Co-teaching}.
\end{abstract}

\section{Introduction} \label{sec::intro}
The availability of large-scale datasets with clean annotations has made indispensable contributions to the prosperity of deep learning.
However, collecting these extensive high-quality data has always been a major challenge since the procedure is both expensive and time-consuming.
Appealed by the inexpensive and convenient accesses to large but  defective data, such as querying commercial search engines \cite{li2017webvision}, downloading images from social media \cite{mahajan2018exploring} and other web crawling strategies \cite{olston2010web}, efforts have been made in literature to learn with imperfect data, among which learning with noisy labels has been attached great importance. 

The problem of learning with noisy labels dates back to \cite{angluin1988learning, quinlan1986induction} and the mainstream solutions include adding regularization \cite{ma2018dimensionality, natarajan2013learning}, estimating the label transition matrix \cite{patrini2017making}, training on selected or reweighted samples \cite{han2018co, jiang2018mentornet, malach2017decoupling, wei2020combating, yu2019does}, label correction \cite{PENCIL_CVPR_2019, zhang2021learning}
and other strategies categorized into the semi-supervised learning genre \cite{ding2018semi, kong2019recycling, li2020dividemix}. 
In addition to works that provide concise and theoretically sound frameworks to combat label noise, there are other existing works devoted to achieving the state-of-the-art performances on benchmark datasets which require an appropriate tuning on multiple hyper-parameters~\cite{li2020dividemix, zhang2021learning}.

With the above in mind, we propose a simple method that combines the regularization and training on selected samples paradigms together to improve state-of-the-art performance on two real-world datasets. 
To be specific, we revisit a compression regularization called Nested Dropout \cite{rippel2014learning},
which was originally proposed to learn ordered feature representations, 
such that it can be used to perform fast information retrieval and adaptive data compression.  
It has also been shown in the paper that the ordered feature representation has a strong connection to the PCA solution, that is, the feature channels can be associated to the eigenvectors of the covariance of input data. 
We find this property 
important in combating label noise since we are then able to conduct a signal-to-noise separation on the learned feature channels. 
We verify this intuition empirically, which shows that Nested Dropout gives rise to a strong baseline for learning with noisy labels. 

To further take full advantage of this compression regularization, we combine it with a widely acknowledged method called Co-teaching~\cite{han2018co}.
This is another strong baseline for learning with noisy labels. The basic idea is that two networks can be trained simultaneously where each network updates itself based on the small-loss mini-batch samples selected by its peer. 
The success of Co-teaching requires the two networks to be reliable enough to select clean samples for each other where we assume the smaller the loss, the cleaner the data \cite{han2018co, jiang2018mentornet, kumar2010self, tanaka2018joint, yu2019does}.
In this regard, we propose a two-stage solution: 
\begin{itemize}
\item In stage one, two Nested Dropout networks are trained separately to provide reliable base networks for the subsequent stage; 
\item In stage two, the two trained networks are further fine-tuned with Co-teaching.
\end{itemize}
As such, we are able to boost the classical strategy with a simple compression regularization.

The rest of the paper is organized as follows: 
Section \ref{sec::method} gives the architecture of the proposed method. 
Section \ref{sec::exp} presents the experiments on an illustrative toy dataset and two real-world datasets, namely, Clothing1M \cite{xiao2015learning} and ANIMAL-10N \cite{song2019selfie}. 
Empirical results demonstrate the effectiveness of our two-stage method given its superior performance comparing to several state-of-the-art approaches such as DivideMix \cite{li2020dividemix} and PLC \cite{zhang2021learning}.

\begin{figure*}[t]
	\begin{minipage}[t]{0.24\textwidth}  
		\centering 
		\addtocounter{figure}{-1}
		\includegraphics[width=\textwidth]{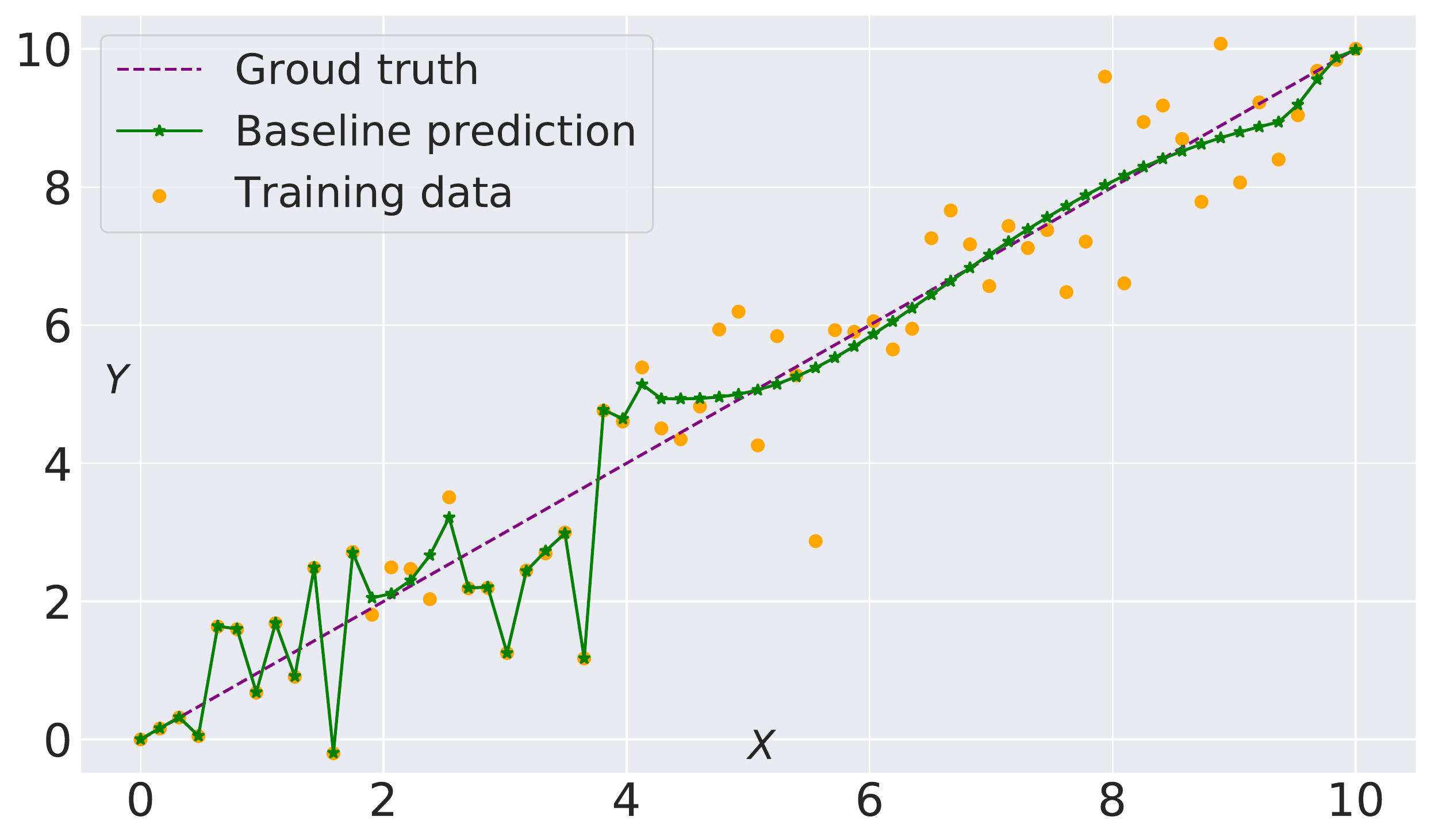}
        \captionsetup{labelformat=empty}
		\caption{(a) MLP}
	\end{minipage}  
	\begin{minipage}[t]{0.24\textwidth}  
		\centering  
		\addtocounter{figure}{-1}
		\includegraphics[width=\textwidth]{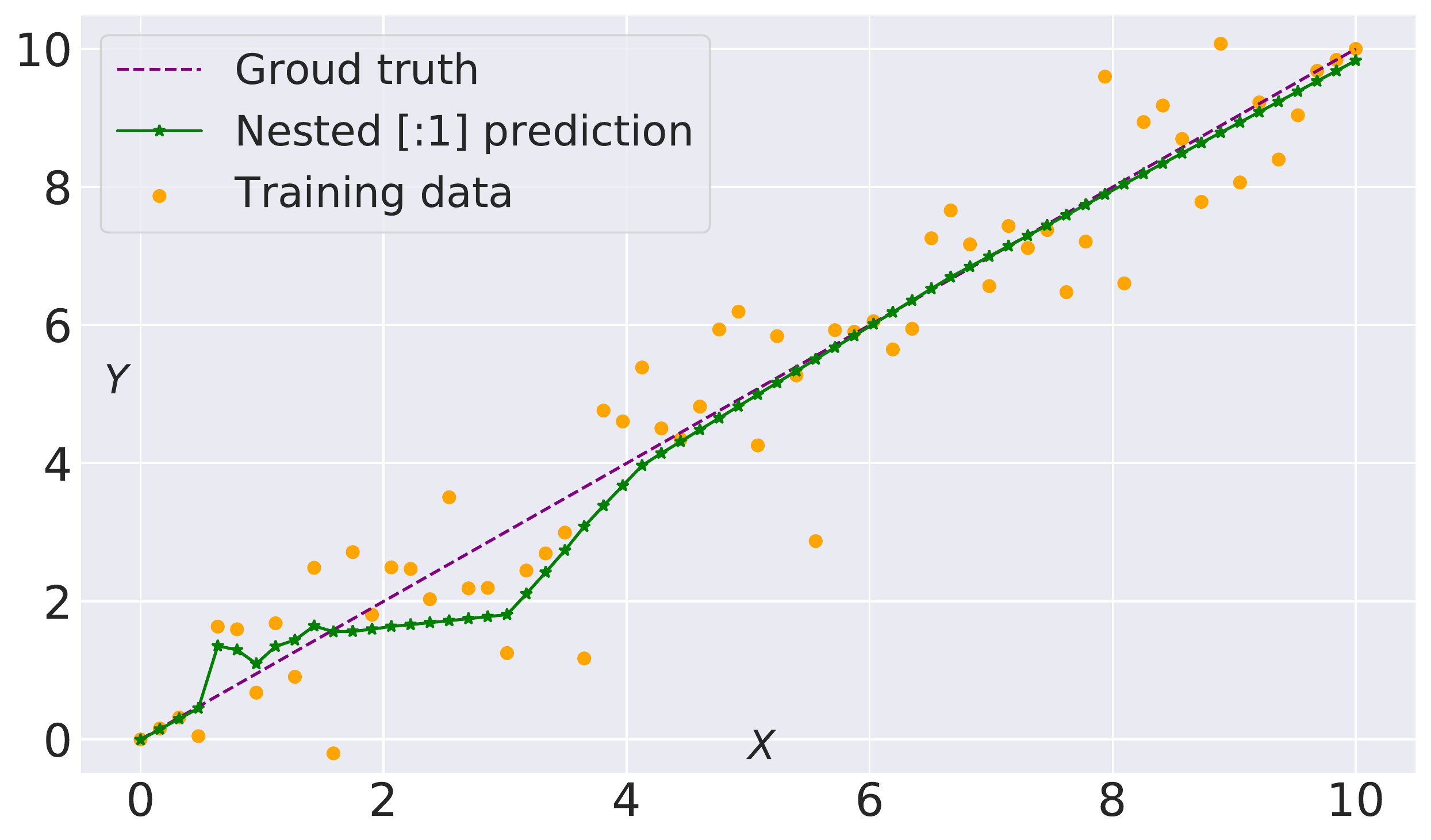} 
		\captionsetup{labelformat=empty}
        \caption{(b) MLP$+$Nested $k$=1}
	\end{minipage} 
	\begin{minipage}[t]{0.24\textwidth}  
		\centering  
		\addtocounter{figure}{-1}
		\includegraphics[width=\textwidth]{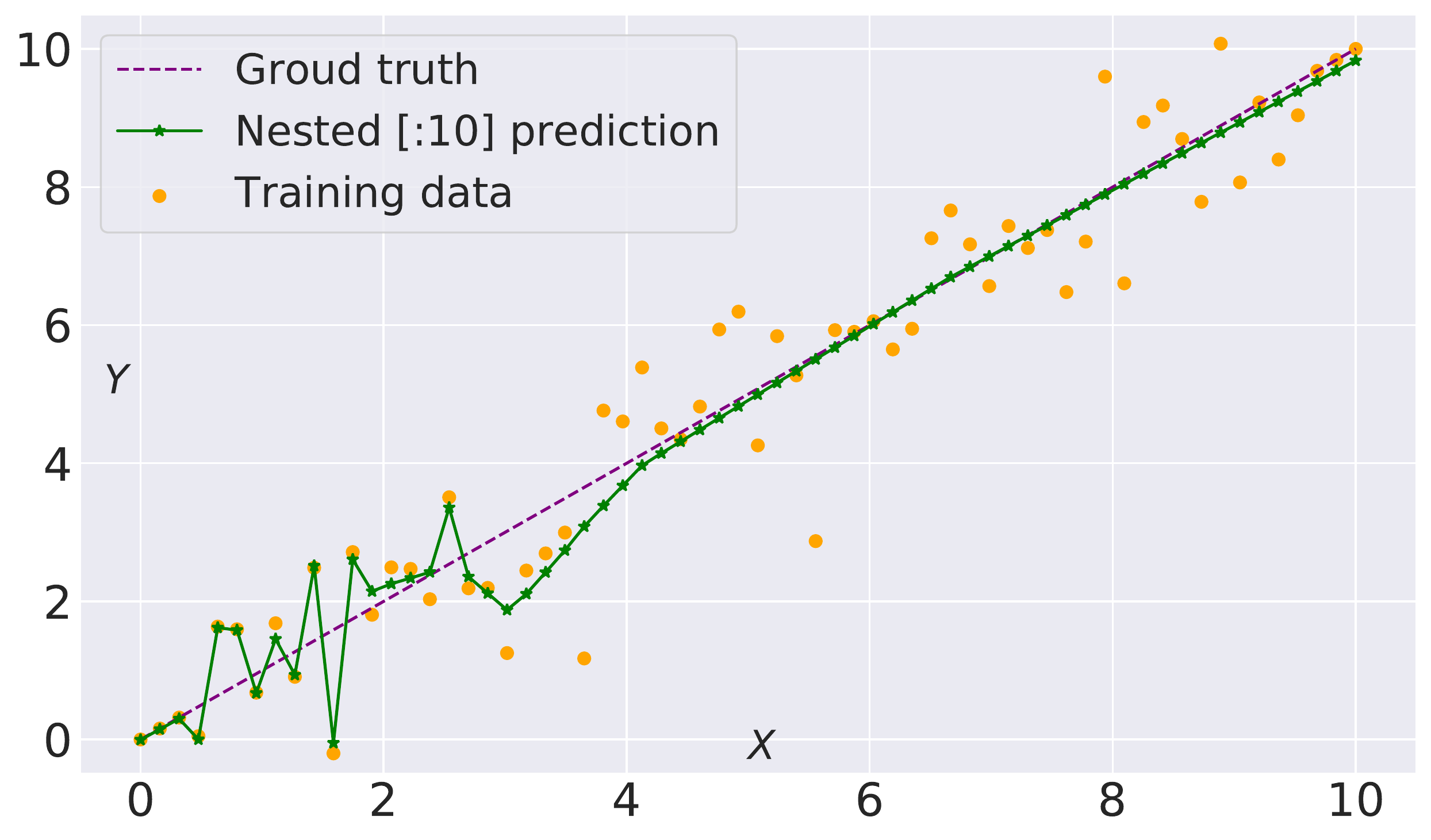} 
		\captionsetup{labelformat=empty}
        \caption{(c) MLP$+$Nested $k$=10}
	\end{minipage} 
	\begin{minipage}[t]{0.24\textwidth}  
		\centering  
		\addtocounter{figure}{-1}
		\includegraphics[width=\textwidth]{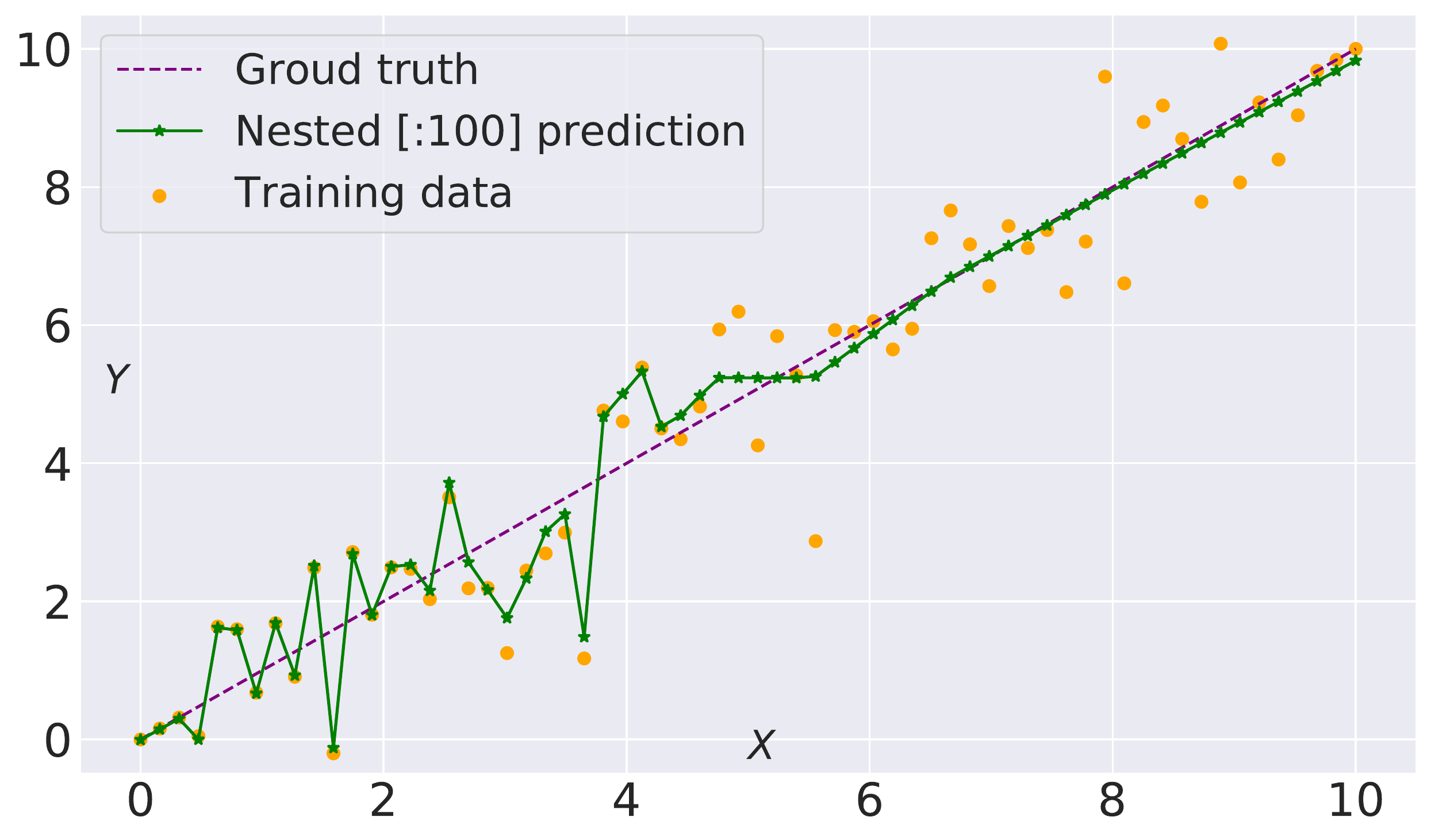}
        \captionsetup{labelformat=empty}
		\caption{(d) MLP$+$Nested $k$=100}
	\end{minipage}  
	\centering  
	\caption{Comparisons of regression between MLP and MLP incorporated with Nested Dropout~\cite{rippel2014learning} on a synthetic noisy label dataset. 
	(a) Training the regression with standard MLP; (b-d) Learning regression with MLP$+$Nested and plot the prediction results using only the first $k$ channels where (b) $k$ = 1, (c) $k$ = 10, (d) $k$ = 100.}
	\label{fig::ResistNoise}
\end{figure*} 

\section{Method} \label{sec::method}
In this section, we present our approach. We find that the compression regularization is a simple yet effective technique to combat label noise. 
Specifically, we first focus on one compression regularization named Nested Dropout~\cite{rippel2014learning} in Section \ref{subsec::nested}, which may serve as a stronger substitute of Dropout~\cite{srivastava2014dropout}.
To take full advantage of Nested Dropout, we further combine it with one commonly accepted method called Co-teaching~\cite{han2018co} in Section \ref{subsec::co_teaching}. 

\subsection{Nested Dropout} \label{subsec::nested}
Nested Dropout~\cite{rippel2014learning} is one regularization able to learn ordered representations where different dimensions have different degrees of importance. 
While applied, meaningless representations can be dropped, leading to a compressed network~\cite{gomez2019learning}.
Taking above into consideration, these ordered representations could be particularly adapted to learning with noisy labels since representations learned from noisy data are expected to be meaningless.

Let $h \in \mathbb{R}^{K \times H \times W}$ be the hidden feature representation obtained by the feature network $f$, i.e.~$h = f(x)$ with $x$ being the input. 
To obtain an ordered feature representation, in each training iteration, we only keep the first $k$ dimensional feature of $h$ and mask the rest to zeros, that is,
\begin{align*}
    z = \big[h_{1:k},\, 0, \ldots, 0 \big] \in \mathbb{R}^{K \times \cdots}
\end{align*}
where $k$ is sampled from a categorized distribution with parameters
\begin{align} 
\label{eq:CatGaussian}
    \Big\{ p_k \propto \exp\Big( -\frac{k^2}{2\, \sigma_{\text{nest}}^2} \Big), \quad \forall k=1,\ldots,K \Big\}.
\end{align}
where $\sigma_{\text{nest}}$ is one major hyper-parameter in our method.
In this manner, smaller $k$'s are preferred when $\sigma_{\text{nest}}$ is small. 





\subsection{Co-teaching} \label{subsec::co_teaching}
Co-teaching~\cite{han2018co} is a standard baseline for learning with label noise.
The idea is to train two deep networks $f_1$ and $f_2$ simultaneously where each network selects its (1 - $\lambda_{\text{forget}}$) percent small-loss instances, i.e.~$\mathcal{D}_1$ and $\mathcal{D}_2$, respectively, where 
$\lambda_{\text{forget}}$ is the forget rate and it is an important hyper-parameter in the Co-teaching architecture.
Networks update themselves based on the sample set selected by their peers.

Considering that small-loss instances are more likely to be clean \cite{han2018co, jiang2018mentornet, kumar2010self, tanaka2018joint, yu2019does},
we could obtain classifiers resistant to noisy labels by training them on these instances. 
However, the above comes with one premise that the classifier should be reliable enough so that the small-loss instances are indeed clean. 
In the original {Co-teaching}~\cite{han2018co}, it proposes to keep all the instances in the mini-batch at the beginning, and then gradually decrease the number of instances in $\mathcal{D}_1$ and $\mathcal{D}_2$ until the $N$-th epoch, after which the number of samples used to train the models becomes fixed.
Compared to tuning the hyper-parameter $N$, we find it more stable to conduct standard training for each model until convergence then fine-tune both models with Co-teaching.


To combine with Nested Dropout, the training procedure comes with two stages:
\textit{(i)} train two Nested Dropout networks separately;
\textit{(ii)} fine-tune these two networks with {Co-teaching}.

In the first stage, we set a learning rate warm-up to cope with the difficulty of training with Nested Dropout in early epochs resulting from the high probability of dropping most of the channels in the feature layer (i.e. large $\sigma_{\text{nest}}$).
In the second stage, Nested Dropout is maintained during the training of each model except for selecting small-loss instances $\mathcal{D}_1$ and $\mathcal{D}_2$. The final performance is the accuracy of the ensembled model.



\section{Experiments} \label{sec::exp}
In this section, we present our experimental results. 
We first show how Nested Dropout deals with regression noise in a toy example in Section~\ref{sec:toy}. 
In Section~\ref{sec:compare}, we compare our method with state-of-the-art approaches on two real-world datasets: Clothing1M~\cite{xiao2015learning} and ANIMAL-10N~\cite{song2019selfie}. 
Finally, an ablation study on ANIMAL-10N~\cite{song2019selfie} is given in Section~\ref{sec::ablation}. 

\subsection{Toy example: a simple regression with noise}
\label{sec:toy}
To gain an intuitive understanding on the reason why Nested Dropout~\cite{rippel2014learning} is able to resist label noise, we present a simulated regression experiment. 
We generate a dataset of noisy observations from $y_i = x_i + \epsilon_i$ for $i=1,\ldots,64$ where $x_i$ is the evenly spaced value between $[0,10]$ and $\epsilon_i \sim \mathcal{N}(0,1)$ are independently sampled. 
We adopt a multilayer perceptron (MLP) composed of three linear layers with input and output dimensions being $1 \rightarrow 64 \rightarrow 128 \rightarrow 1$.
Each layer is followed by a ReLU activation except the last one. 
For the model with Nested Dropout~\cite{rippel2014learning}, we apply
\begin{table}[h]
\caption{Test set accuracy (\%) on Clothing1M~\cite{xiao2015learning}. We report average accuracy as well as the standard deviation for three runs. Results with ``*" are either using a balanced subset or a balanced loss.}
\vspace{-5mm}
\label{tab::clothing1m}
\begin{center}
\begin{tabular}{c|c}
\hline \hline
Methods  & Acc. (\%)   \\ \hline
CE~\cite{wei2020combating} & 67.2 \\ 
F-correction~\cite{wei2020combating,patrini2017making} & 68.9 \\
Decoupling~\cite{wei2020combating,malach2017decoupling} & 68.5 \\
Co-teaching~\cite{wei2020combating,han2018co} & 69.2 \\
Co-teaching+~\cite{wei2020combating,yu2019does} & 59.3 \\
JoCoR~\cite{wei2020combating} & 70.3 \\
JO~\cite{tanaka2018joint} & 72.2 \\
Dropout*~\cite{srivastava2014dropout} & 72.8\\
PENCIL*~\cite{PENCIL_CVPR_2019} & 73.5\\
MLNT~\cite{li2019learning} & 73.5 \\ 
PLC*~\cite{zhang2021learning} & 74.0\\
DivideMix*~\cite{li2020dividemix} & 74.8 \\
\hline
\multicolumn{2}{c}{\textbf{Ours}} \\
Nested* & 73.1 {$\pm$} 0.3\\
Nested + Co-teaching* & \bf 74.9 {$\pm$} 0.2\\
\hline \hline
\end{tabular}
\end{center}
\vspace{-5mm}
\end{table}
Nested Dropout~\cite{rippel2014learning} to the last layer of this MLP and denote it by MLP$+$Nested. 
We set $\sigma_{\text{nest}} = 200$ and use the \eqref{eq:CatGaussian} to sample the number of features for training. 
Results after $100$k epochs are shown in Figure \ref{fig::ResistNoise}.
It can be seen that MLP overfits to the label noise while MLP$+$Nested with the first $k \in \{1, 10\}$ channels recovers the ground-truth $y=x$ better. 
However, with the number of channels increasing, MLP$+$Nested gradually overfits to the label noise due to over parameterization. 
This demonstrates that the first few channels in MLP$+$Nested contain the main data structure information, while channels towards the end are more likely to encode misled information that would overfit to the noise.

\subsection{Comparison with state-of-the-art methods on real datasets}
\label{sec:compare}
\paragraph{Datasets} We conduct experiments on two real datasets: Clothing1M~\cite{xiao2015learning} and ANIMAL-10N~\cite{song2019selfie}. 
The Clothing1M~\cite{xiao2015learning} dataset contains $1$ million clothing images obtained from online shopping websites with $14$ categories. 
The labels in this dataset are quite noisy with an unknown underlying structure. 
This dataset also provides $50$k, $14$k and $10$k manually verified clean data for training, validation and testing, respectively. Note that the clean training set is not used during the training. Following~\cite{PENCIL_CVPR_2019,zhang2021learning}, in our experiment, we randomly sample a balanced subset which includes $260$k images (18.5k images per category) from the noisy training set and use it as our training set and report the classification accuracy on the $10$k clean test data. 
We adopt the standard data augmentation procedures to train ImageNet~\cite{deng2009imagenet,he2016deep}, including random horizontal flip, and resizing the image with a short edge of $256$ and then randomly cropping a $224 \times 224$ patch from the resized image. 
The ANIMAL-10N is recently proposed by~\cite{song2019selfie}. 
It contains $10$ animals with confusing appearance. The estimated label noise rate is $8\%$. There are $50$k training and $5$k testing images. 
We did not apply any data augmentation so that the setting is the same with \cite{song2019selfie}.

\begin{table}[t]
\caption{Test set accuracy (\%) on ANIMAL-10N~\cite{song2019selfie}. We report average accuracy as well as the standard deviation for three runs.}
\vspace{-5mm}
\label{tab::animal}
\begin{center}
\begin{tabular}{c|c}
\hline \hline
Methods  & Acc. (\%)   \\ \hline
CE~\cite{song2019selfie} & 79.4 {$\pm$} 0.1\\
Dropout~\cite{srivastava2014dropout} & 81.3 {$\pm$} 0.3\\
SELFIE~\cite{song2019selfie} & 81.8 {$\pm$} 0.1\\
PLC~\cite{zhang2021learning} & 83.4 {$\pm$} 0.4\\
\hline
\multicolumn{2}{c}{\textbf{Ours}} \\
Nested & 81.3 {$\pm$} 0.6\\
Nested + Co-teaching & \bf 84.1 {$\pm$} 0.1\\
\hline \hline
\end{tabular}
\end{center}
\vspace{-5mm}
\end{table}

\begin{table*}[ht]
\centering
\caption{Average test accuracy (\%) with standard deviation (three runs) of different $\sigma_{\text{nest}}$ on ANIMAL-10N~\cite{song2019selfie}. 
The corresponding optimal number of channels $k^*$ for each model is also provided (entry ``$k^*$"). 
We report test accuracy of single model (entry ``Acc.") as well as the accuracy with the combination of Co-teaching (entry ``Co-teaching Acc.") }
\vspace{-5mm}
\label{tab::ablation}
\begin{center}
\begin{tabular}{c|cc|cc}
\hline \hline
$\sigma_{\text{nest}}$& $k^*$ & Acc. (\%) & $k^*$ & Co-teaching Acc. (\%)  \\ \hline
CE &  4096& 79.4 {$\pm$} 0.1 & 4096 & 82.2 {$\pm$} 1.1 \\
\hline
25
&  17.7 {$\pm$} 9.7 & 81.0 {$\pm$} 0.6 &  16.3 {$\pm$} 6.9 & 83.7 {$\pm$} 0.1 \\ \hline 

50 & 18.8 {$\pm$} 6.9 & \bf 81.3 {$\pm$} 0.6 & 13.4 {$\pm$} 4.1 & 84.1 {$\pm$} 0.2 \\ \hline

100 & 13.6 {$\pm$} 5.6 & 81.0 {$\pm$} 0.5 & 16.8 {$\pm$} 7.1 & \bf 84.1 {$\pm$} 0.1 \\ \hline 

150 & 16.0 {$\pm$} 3.6 &  81.1 {$\pm$} 0.5 & 18.8 {$\pm$} 7.4 &  83.8 {$\pm$} 0.2 \\ \hline 
250 & 13.2 {$\pm$} 3.1 &  81.1 {$\pm$} 0.2 & 21.0 {$\pm$} 10.4  &  83.8 {$\pm$} 0.1 \\
\hline\hline
\end{tabular}%
\end{center}
\vspace{-5mm}
\end{table*}

\paragraph{Implementation details}
We implement our approach on Pytorch. Experiments on Clothing1M~\cite{xiao2015learning} are with ResNet-18~\cite{he2016deep} pre-trained on ImageNet~\cite{deng2009imagenet} following \cite{wei2020combating}. 
Note that Nested Dropout or Dropout is applied right before the linear classifier in the network.
Models in stage one are optimised with SGD optimizer with a momentum of $0.9$, a weight decay of $5e{-4}$, an initial learning rate of $2e{-2}$, and batch size of $448$. 
During training, we run learning rate warm-up for $6000$ iterations, then train the model for $30$ epochs with the learning rate decayed by $0.1$ after the $5$th epoch. 
In stage two, we apply Co-teaching to fine-tune two well-trained models. SGD optimizer is utilized with the same settings only with an initial learning rate changing to $2e{-3}$. 
Moreover, we set 
forget rate $\lambda_{\text{forget}}$ in the Co-teaching~\cite{han2018co} to be $0.3$, freeze batch norm and no warm-up is applied. Models are again trained for $30$ epochs with the learning rate decayed by $0.1$ after the $5$th epoch.

For ANIMAL-10N, we use VGG-19~\cite{simonyan2014very} with batch normalization \cite{ioffe2015batch} as in \cite{song2019selfie}. 
The two Dropout layers in the original VGG-19 architecture are changed to Nested Dropouts when Nested is applied.
The SGD optimizer is employed. Following~\cite{song2019selfie}, we train the network for $100$ epochs and use an initial learning rate of $0.1$, which is divided by $5$ at $50\%$ and $75\%$ of the total number of epochs. In stage one, models are trained with learning rate warm-up for $6000$ iterations. 
In stage two, no warm-up is applied, batch norms are freezed, forget rate $\lambda_{\text{forget}}$ is $0.2$, initial learning rate is $4e{-3}$ and decayed by $0.2$ after the $5$th epoch with $30$ epochs in total. 

\vspace{-3mm}
\paragraph{Results on the Clothing1M~\cite{xiao2015learning}} 
We now compare our method to state-of-the-art approaches on Clothing1M~\cite{xiao2015learning} in Table~\ref{tab::clothing1m}. 
It is worth noting that Table~\ref{tab::clothing1m} also includes very recent approaches such as DivideMix~\cite{li2020dividemix} and PLC~\cite{zhang2021learning}. 
Surprisingly, our single model with Nested Dropout~\cite{rippel2014learning} not only surpasses the standard Dropout~\cite{srivastava2014dropout}, but also achieves comparable performance to PENCIL~\cite{PENCIL_CVPR_2019} and MLNT~\cite{PENCIL_CVPR_2019}. 
Incorporating Co-teaching~\cite{han2018co} further boosts the performance to $74.9\%$ and outperforms the state-of-the-art DivideMix~\cite{li2020dividemix}. 

\vspace{-2mm}
\paragraph{Results on the ANIMAL-10N~\cite{xiao2015learning}} 
Experimental results on ANIMAL-10N~\cite{xiao2015learning} are given in Table~\ref{tab::animal}. 
The dataset is recently proposed, and we compare with two approaches that report performance on this dataset: SELFIE~\cite{song2019selfie} and  PLC~\cite{zhang2021learning}. 
It can be seen that our single model achieves comparable performance to Dropout as well as SELFIE~\cite{song2019selfie} and Co-teaching provides a consistent performance boost, which is similar to the results on the Clothing1M~\cite{xiao2015learning}. 
Note that, our best performance by using Nested Dropout~\cite{rippel2014learning} and Co-teaching~\cite{han2018co} achieves $84.1\%$ accuracy outperforms recent approach PLC~\cite{zhang2021learning} by $0.7\%$. 

\subsection{Ablation study}
\label{sec::ablation}
In this section, we provide the ablation study of $\sigma_{\text{nest}}$ on ANIMAL-10N~\cite{song2019selfie}. 
The results are given in Table~\ref{tab::ablation}. 
As we can see, the Nested Dropout~\cite{rippel2014learning} provides consistent improvement compared to training with standard cross entropy loss (entry ``CE") and the performance gain is also robust to the choices of the hyper-parameter $\sigma_{\text{nest}}$. 
Moreover, fine-tuning with Co-teaching~\cite{han2018co} provides clear boost for all the models. 
We also show the optimal number of channels of each model (entry ``$k^*$") in the table. 
Note that though two layers of Nested Dropout have been applied to the classifier of VGG-19, the optimal number of channels $k^*$ is recorded with regard to the last Nested Dropout layer for simplicity.
Interestingly, the models trained with Nested Dropout~\cite{rippel2014learning} achieve better performance but with only using less than 1\% of channels compare to the models trained with standard cross entropy (entry ``CE").

\section{Conclusion}
In this paper, we investigated the problem of image classification in the presence of noisy labels.
Specifically, we first demonstrated that a simple compression regularization called Nested Dropout~\cite{rippel2014learning}
can be used to combat label noise.
Moreover, due to its simplicity, Nested Dropout~\cite{rippel2014learning} can be easy combined with Co-teaching~\cite{han2018co} to further boost the performance. We validated our approach on two real-world noisy datasets and achieved state-of-the-art performance on both datasets. The proposed approach is simple comparing to many existing methods. 
Therefore, we hope that our approach can be served as a strong baseline for future research on learning with noisy label.

\vspace{-2mm}
\paragraph{Acknowledgements} The research leading to these results has received funding from the European Research Council under the European Union's Horizon 2020 research and innovation program / ERC Advanced Grant E-DUALITY (787960).
This paper reflects only the authors' views and the Union is not liable for any use that may be made of the contained information. 
This work was also supported in part by 
EU H2020 ICT-48 Network TAILOR (Foundations of Trustworthy
AI - Integrating Reasoning, Learning and Optimization),
Leuven.AI Institute.


{\small
\bibliographystyle{ieee_fullname}
\bibliography{egbib}
}

\end{document}